# Correct classification for big/smart/fast data machine learning

Sander Stepanov

sander.stepanov@gmail.com

## Abstract

Table (database) / Relational database Classification for big/smart/fast data machine learning is one of the most important tasks of predictive analytics and extracting valuable information from data. It is core applied technique for what now understood under data science and/or artificial intelligence. Widely used Decision Tree (Random Forest) and rare used rule based PRISM, VFST, etc classifiers are empirical substitutions of theoretically correct to use Boolean functions minimization. Developing Minimization of Boolean functions algorithms is started long time ago by Edward Veitch's 1952. Since it, big efforts by wide scientific/industrial community was done to find feasible solution of Boolean functions minimization. In this paper we propose consider table data classification from mathematical point of view, as minimization of Boolean functions. It is shown that data representation may be transformed to Boolean functions form and how to use known algorithms. For simplicity, binary output function is used for development, what opens doors for multivalued outputs developments.

## I. Introduction

Many practical applications of machine learning deals with statistical inference about data represented by table, where rows are observations and last column is symbolic encoding of object related to these observations. Logical model for data implies statistical/random relation between observations and objects, when sometime the same observations may be observed for different objects. But generally, there is some statistical connection between observed data and objects. Very often in practical cases combinatorial analysis/search is not possible due to limited resources of computation facilities, then they say data is too big for computing resources. Remarkable is slide number 9 from [Zhu] on Fig. 1

B. Efron (2010), Large-Scale Inference: Empirical Bayes Methods for Estimation, Testing, and Prediction, Cambridge University Press.

- testing thousands of hypotheses at the same time

- e.g., 1000 hypotheses, each tested

at significance level of 0.05 ⇒ expect to find 50 "significant" hypotheses just

by chance, even if none of them is

- not so much machine learning, but definitely big data

Fig 1, slide number 9 from [Zhu]

It means, when observation have many features, then even not existing rules of relation between observations and objects may be found as existing. Especially, in practice often non observed data needs to be classified basing on observed data from past. Axiomatic approach is to assume that observed data has some features values, those leads to object observation.

For example in context of what usually understood as big data, lets say some data table has 1000 features, features have values or 1 or 0 and number of observations is 10 000 000, when:

for features with numbers from 1 to 5 , observed values equal to 1 lead to object of class 1, when number of such kind of observations is 2000 ( what is something small in comparison to overall number of observations) and never lead to object of class 0 , ;

for features with numbers from 8 to 1 , observed values equal to 1 lead to object of class 1 when number of such kind of observations is 3000 ( what is something small in comparison to overall number of observations) and never lead to object of class 0.

Other several features has similar very complicated to find by combinatorial search relation between features values and objects. But vast majority of features are not related at all to objects, sometime they called as dummy features.

Now lets randomly permutated features, it is clear by combinatorial search these relevant features may not be found.

Usually, vey simplified data is assumed to be, when some features have simple causality, for example when mutual information between some feature and object has not small value, for example for someone particular feature conditional probability to see given class when this feature has some particular value which is not small. Then sequential splitting observed data is used: well known Decision Trees , PRISM [Cendrowska] , VFDR [Gama] , IEBRG [Gegov] , Hidden Decision Trees [Granville] .

In such kind of not easy, but realistic,   data case it is reasonable to assume of existence of simple and as simple as possible logical function for observations and objects.   When this actual underlying relations is more complicated, we will get simpler rule, what actually even better.

Regarding to data size to call it big data, data structure may cause data to be combinatorically impossible to analyse, even for data size considered now as small data [Carlsson]. There for Sagested approach is applicable to much more cases, then complicated data with large sizes.

To sum up,   minimization of Boolean functions is proper mathematics to use for binary data. Relating to complexity of calculations:

Unbalanced data lead to reduction of calculations;

Nature of Boolean functions gives opportunity to use distributed computing: data maybe divided to parts, each part calculated independently, each part minimized Boolean functions are processed to get global minimization;

Real time calculation is accomplished by minimisation of result of minimisation of previous data and new data. It is good question, how to weight past data observations like exponential mean technique.

## II. Observations transformation to binary format

There are two options to transform features to binary format: one hot encoding and binary form of continues/integer values. One hot encoding for categorical variable introduce redundancy and makes data matrix to be sparse. It is subject to further research how sparsity and redundancy influence on calculation complexity and resulting model. Discretization of continues values [Discretization of continuous features] followed by continues number transformation to binary representation is well known operation. It is art, which may be done in many ways, but simple approach may be to choose number of levels and encode levels to binary format.

## III. Boolean Functions Minimization

Boolean encoded features minimization is extensively developed area for many years, so there is no needs to develop entirely new numeric methods, therefor using proposed approach is correct industrial engineering design [Seth Godin]. Here we mention widely known Quine–McCluskey algorithm [Quine–McCluskey] and its calculation reduction Espresso minimizer [Espresso]. Author has some reservation about comparison of suggested method with RETE [Carole-Ann Berlioz] , [Gupta] , [Miranker] and RETE inspired methods. But for example some inequality rule x < 4 then 0 , else 1; where 4 is integer number and for x in range from 0 to 15 , rule may be encoded by Boolean function. 16 integer number may be encoded by 4 digit binary representation, so number 3 will be 0011, lets write it in form b4 =0, b3 =0, b2=1, b3=1, then rule will be  "or b4 or b3".

Described approach is pure scientific approach ( here we build mathematical model from real word data and then show existing well developed by many people during many years tools to find solution for this mathematical model ). There is mentioning Quine-McCluskey Boolean minimization for big data context of association rules reduction in [Seol], for purpose of finding important association rules some data is removed. Conversely in classification case, these removed rules are need to be not removed, since these rare observations are very possible to be causal for target classes.  Suggested approach is very similar to using Boolean minimization for Configurational comparative methods [Thiem], but different in application area.

There is a lot to do from data science prospective – to adjust real word data to SAT formulation [Villa] , [Cordone] , [Kagliwal] , [Coudert] , [Roy] , [Goldberg ] , [Xiao Yu Li ] , [FLoC Olympic Games] , [Satlive]  , [Max-SAT Evaluation] , [Satcompetition] , [SAT-Race]. Not all data is observed, therefor there is question how to consider non observed data, to which class make assumption is belongs to. In exemplary case of 100 features, there are 2 in power 100 observation possible, but if for data table of 10 million observation we have imbalanced data, that   0.1% of data describes class 1. Then there are several ways to consider unobserved data, for example make decision to put it in "do not care", or just to find SAT solution as is, etc. Some bagging may be considered, too: to find solution not for all available data but for several data parts. Then find general rules.

In addition to SAT and generating correlated Gaussian data [scikit-learn] , [Alice], another recent breakthrough enables considered approach to use is synthetic data generation [GenOrd] , [OrdNor] , [datasets over algorithms ] , [Ratner]. Till now mainly was approach to use real data for machine learning development.

IV. Interpretability and Causality vs Black Box Model vs The Hardest Part of Data Science (Yanir Seroussi)

Lets consider example, given data table as described in introduction. Then for features with numbers from 1 to 5 with 2000 observation casually related to observed class 1 ,    there may be many observations with the same values generated randomly for both class 1 and class 0  - something like generative randomness per [Cooper]. Therefore, it is data scientist who is the magician to make this "magic"  mentioned in    "magic is to build an in-depth understanding of the domain knowledge and available data assets"[Seroussi] and " Data science projects vary from "executive dashboards" through "automate what my analysts are already doing well" to "here is some data, we would like some magic." " [win-vector] and make results to be logical [Ribeiro] and we can trust them [bobg].

There are attempts to preserve more observed data information, rather than distilling classification rules from any redundant information.

R package CNA describes this on page 2: " it does not eliminate redundancies from sufficient and necessary conditions by means of Quine-McCluskey optimization (Quine 1959, McCluskey 1965), but by means of an optimization algorithm that is custom-built for causal modeling"

Another idea to save information about similar variables presented in [Kim].

V. Conclusion

This paper draws attention to SAT use in Boolean data tables classification cases for machine learning. Rapidly developed SAT solutions may be magic tool for  big/smart/fast data applications. Gradually [The Emperor's New Clothes ]  classification of big data moving from practicable decisions of theoretically relatively simple infrastructure problems [Stucchio] , [Pafka] , [Fowler], [Appuswamy], [Mims]. Soon as applications software became unbuggy [Houg] and will be enable provide running machine learning algorithms on many computers in parallel (not separate parts for different of algorithms, but whole algorithms – what databricks.com calls "spark as scheduler" [From Quick Start to Scikit-Learn] , [spark-sklearn] ) and machine learning use on many computers become unmessy [Matthies] then algorithmically part start to be appreciated and this time is very close.

https://cran.r-project.org/web/packages/QCA/index.html

[Thiem] Thiem, Alrik and Michael Baumgartner. 2016. Glossary for Configurational Comparative

Methods, Version 1.1. In: Thiem, Alrik. QCApro: Professional Functionality for

Performing and Evaluating Qualitative Comparative Analysis, R Package Version 1.1-1.

URL: http://www.alrik-thiem.net/software/.

[Thiem]  Michael Baumgartner and Alrik Thiem

Identifying Complex Causal Dependencies in Configurational Data with Coincidence Analysis

The R Journal Vol. 7/1, June 2015

https://journal.r-project.org/archive/2015-1/baumgartner-thiem.pdf

[Thiem]   Ambuehl, Mathias, Michael Baumgartner, Ruedi Epple, Alexis Kauffmann and Alrik Thiem

cna: A Package for Coincidence Analysis (CNA)

https://cran.r-project.org/web/packages/cna/

https://cran.r-project.org/web/packages/cna/cna.pdf

page 2 it does not eliminate redundancies from sufficient and necessary conditions by means of

Quine-McCluskey optimization (Quine 1959, McCluskey 1965), but by means of an optimization

algorithm that is custom-built for causal modeling

[Cooper] Barry Cooper, Judith Glaesser

The set theoretic analysis of probabilistic regularities with fsQCA, QCApro and

CNA: Exploring possible implications for current "best practice", 2016

http://www.compasss.org/

[Seroussi]  Yanir Seroussi  The hardest part of data science

https://yanirseroussi.com/2015/11/23/the-hardest-parts-of-data-science/

[Seroussi]  Ask why! Finding motives, causes, and purpose in data science

https://yanirseroussi.com/2016/09/19/ask-why-finding-motives-causes-and-purpose-in-data-science/

https://www.youtube.com/watch?v=2wqu-drqlpo

8  Correct classification for big smart fast data machine learning  Sep27  2016 Sander Stepanov

11  Correct classification for big smart fast data machine learning  Sep27 2016 Sander Stepanov

12  Correct classification for big smart fast data machine learning  Sep27  2016 Sander Stepanov

http://minisat.se/MiniSat.html

http://minisat.se/SatELite.html

SatELite is a CNF minimizer, intended to be used as a preprocessor to the SAT solver. It is designed to compress the CNF fast enough not to be a bottle neck,

[Samir Sapra ] Samir Sapra ,  Michael Theobald , Edmund Clarke SAT-Based Algorithms for Logic Minimization, Computer Design, 2003. Proceedings. 21st International Conference
http://www.cs.cmu.edu/~emc/papers/Conference%20Papers/SAT-Based%20Algorithms%20for%20Logic%20Minimization.pdf

http://ieeexplore.ieee.org/document/1240948/

This section presents new SAT-based implementations of the REDUCE operator. In particular, we describe three methods in order of increasing performance advantage over the operator implementation in ESPRESSO-II. Our best implementation outperforms ESPRESSO-II's REDUCE by

more than a factor of 100 on many of our large examples

[Sapra] Samir Sapra , Using SAT Checkers to Solve The Logic Minimization Problem

, Thesis, School of Computer Science  Carnegie Mellon University , Advisor: Edmund Clarke , Post-Doc Advisor: Michael Theobald , May 3, 2003

http://www.cs.cmu.edu/afs/cs/user/mjs/ftp/thesis-program/2003/sapra.pdf

[Carlsson] Gunnar Carlsson: The Shape of Data (DARPA "Wait, What?")

https://www.youtube.com/watch?v=X9ktWgJ7ung&feature=youtu.be

[spark-sklearn] Scikit-learn integration package for Apache Spark

https://github.com/databricks/spark-sklearn

[Hunter] Tim Hunter, Combining Machine Learning Frameworks with Apache Spark, spark as scheduler 8 min

https://www.youtube.com/watch?v=yXjQLCu0UMY

Apache® Spark™ MLlib: From Quick Start to Scikit-Learn , spark as scheduler

http://go.databricks.com/spark-mllib-from-quick-start-to-scikit-learn

[Matthies] Anne Matthies - Zero-Administration Data Pipelines using AWS Simple Workflow

https://www.youtube.com/watch?v=RiONy5W9Afk

[Houg] Juliet Houg, Best Practices for running PySpark

https://www.youtube.com/watch?v=cpEOV0GhiHU

[Triantaphyllou] Triantaphyllou, Data Mining and Knowledge Discovery via Logic-Based Methods Theory, Algorithms, and Applications, Springer, 2010

[bobg] bobg, I need an AI BS-Meter

https://gab41.lab41.org/i-need-an-ai-bs-meter-27e94d48c8c1#.2vcy0lzc0

[Ratner] Alexander Ratner, Christopher De Sa, Sen Wu, Daniel Selsam, Christopher Ré, Data Programming: Creating Large Training Sets, Quickly

https://arxiv.org/abs/1605.07723

[datasets over algorithms] datasets over algorithms

http://www.spacemachine.net/views/2016/3/datasets-over-algorithms

[OrdNor] OrdNor: Concurrent Generation of Ordinal and Normal Data with Given Correlation Matrix and Marginal Distributions

https://cran.r-project.org/web/packages/OrdNor/index.html

[GenOrd] GenOrd: Simulation of Discrete Random Variables with Given Correlation Matrix and Marginal Distributions

https://cran.r-project.org/web/packages/GenOrd/

[scikit-learn] Simulate regression data with a correlated design from Sparse recovery: feature selection for sparse linear models

http://scikit-learn.org/stable/auto_examples/linear_model/plot_sparse_recovery.html#example-linear-model-plot-sparse-recovery-py

15  Correct classification for big smart fast data machine learning  Sep27  2016 Sander Stepanov